\documentclass{bmvc2k}

\usepackage{graphics} 
\usepackage{epsfig} 
\usepackage{mathptmx} 
\usepackage{times} 
\usepackage{amsmath} 
\usepackage{amssymb}  
\usepackage{subfigure}
\usepackage{graphicx}
\usepackage{adjustbox}
\usepackage{hhline}
\usepackage{tabularx}
\usepackage{verbatim}
\usepackage{enumitem}

\usepackage{mathtools}

\usepackage[T1]{fontenc}
\usepackage{mathpazo}

\usepackage{extarrows}
\newcommand{\eqdef}{\xlongequal{\text{$\norm{W_j}=1$}}}%

\usepackage{amsmath}
\newcommand\numeq[1]%
  {\stackrel{\scriptscriptstyle \mkern-1.5mu#1 \mkern-1.5mu}{\eqdef}}

\DeclarePairedDelimiter{\norm}{\lVert}{\rVert}

\title{Expression, Affect, Action Unit Recognition: Aff-Wild2, Multi-Task Learning and ArcFace}

\addauthor{Dimitrios Kollias$^{1,2}$}{dimitrios.kollias15@imperial.ac.uk}{1}
\addauthor{Stefanos Zafeiriou$^{1,2}$}{s.zafeiriou@imperial.ac.uk}{2}

\addinstitution{$^{1}$
Department of Computing\\
Imperial College London\\
London, UK
}
\addinstitution{$^{2}$ 
FaceSoft.io
}

\runninghead{Kollias, Zafeiriou}{Aff-Wild2, Multi-Task Learning \& ArcFace}

\begin{document}

\maketitle

\begin{abstract}
Affective computing has been largely limited in terms of available data resources. 
The need to collect and annotate diverse in-the-wild datasets has become apparent with the rise of deep learning models, as the default approach to address any computer vision task.
Some in-the-wild databases have been recently proposed. However: i) their size is small, ii) they are not audiovisual, iii) only a small part is manually annotated, iv) they  contain a small number of subjects, or v) they are not annotated for all main behavior tasks (valence-arousal estimation, action unit detection and basic expression classification). To address these, we substantially extend the largest available in-the-wild database (Aff-Wild) to study continuous emotions such as valence and arousal. Furthermore, we annotate parts of the database with basic expressions and action units. As a consequence, for the first time, this allows the joint study of all three types of behavior states. We call this database Aff-Wild2. We conduct extensive experiments with CNN and CNN-RNN architectures that use visual and audio modalities; these networks are trained on Aff-Wild2 and their performance is then evaluated on 10 publicly available emotion databases. We show that the networks achieve state-of-the-art performance for the emotion recognition tasks. Additionally, we adapt the ArcFace loss function in the emotion recognition context and use it for training two new networks on Aff-Wild2 and then re-train them in a variety of diverse expression recognition databases. The networks are shown to improve the existing state-of-the-art. The database, emotion recognition models and source code are available at \url{http://ibug.doc.ic.ac.uk/resources/aff-wild2}.
\end{abstract}

\section{Introduction}
\label{sec:intro}

Until recently affective computing has been mostly studied in controlled settings of the environments \cite{gross2010multi,lucey2010extended,tian2001recognizing,yin20063d,yin2008high}, with limited amount of participants \cite{pantic2005web,valstar2010induced,lyons1998japanese,ringeval2013introducing,bilakhia2015mahnob}, using pre-defined scenarios that users have to follow, depicting posed expressions \cite{aifanti2010mug,mckeown2011semaine}. 
However, with the development of large and diverse datasets in the field of computer vision (and the accompanying performance gains), it has become apparent that the diversity of human participants and spontaneous expressions have to become the prerogatives in deployment of the affective computing models in practice. 
Large datasets with in-the-wild settings have been recently collected  to study facial expression (Expr) analysis \cite{dalgleish2000handbook,cowie2003describing}, facial action units (AUs) \cite{ekman2002facial} and continuous emotions of valence and arousal (VA) \cite{whissel1989dictionary,russell1978evidence} in-the-wild. 

In \cite{mollahosseini2017affectnet} a static in-the-wild database (AffectNet) has been created that contains VA annotations for $1M$ images. However, from those images, only around $450K$ are manually annotated and from those only $350K$ are valid faces. This number is moderate for training deep neural networks (DNNs). Also this database is static, meaning that it contains only images and no video/audio. It is worth to mention that this database also contains annotations for the seven basic expressions plus the contempt class for $290K$ images.
In \cite{emotionet2016} a static in-the-wild database (Emotionet) has been created that contains AU annotations for $1M$ images. However, from those images, only $50K$ are manually annotated. Half of those consist the validation set and the other half the test set. Again, these sets have a small size to adequately train DNNs and generalize on other databases. Emotionet database also contains annotations for 6 basic and 10 compound emotion categories. Nevertheless, their total size is less than $3K$ and the classes are heavily imbalanced, making it impossible to train DNNs.
In \cite{kollias2019deep,zafeiriou2017aff}, the authors have developed the largest existing audiovisual (A/V) in-the-wild database annotated in terms of VA for around $1.25M$. All these annotations are manual. However this database contains annotations only for VA and the number of subjects in the videos is moderate (298 subjects in total).

Up to the present, there is no database that contains annotations for all main behavior tasks (VA estimation, AU detection, Expr classification). Also, most of the existing databases do not contain sufficiently large numbers of annotated samples for effectively training DNNs. The first contribution in this paper is the creation of a new dataset that contains \textit{260} videos with around \textit{1.4M} frames, annotated for VA. We merge this dataset with Aff-Wild (since this database also contains annotated videos), generating the so-called Aff-Wild2 database. Next, we annotate parts of Aff-Wild2 with AUs and seven basic expression labels, creating about \textit{398K} and \textit{403K} AU and Expr annotations, respectively. To the best of our knowledge, Aff-Wild2 is the first  large scale in-the-wild database containing annotations for all 3 main behavior tasks. It is also the first audiovisual database with annotations for AUs. All AU annotated databases do not contain audio, but only images or videos. 

Next, we conduct multi-task experiments on this database, for emotion recognition. Many questions arise hereafter: how can we combine these three tasks? what loss function should we use? The first apparent answer is to use a loss function equal to the sum of the loss functions of each task. The binary cross entropy loss is used for AU detection. The MSE and Concordance Correlation Coefficient (CCC) losses are used for VA estimation. The standard loss for expression classification is the categorical cross entropy. 
The second contribution of the paper is the development of multi-task CNNs, multi-task CNN-RNNs and  multi-modal, multi-task CNN-RNNs, which are trained on Aff-Wild2 and then applied to \textit{10} publicly available databases (including the Aff-Wild one). The results are very promising, beating the state-of-the-art on emotion recognition in these databases; exceptions are two databases annotated for Expr Recognition. In one of them, a best performing network used a locality preserving loss function \cite{li2017reliable}. This, as well as the recent tendency to develop elaborate loss functions for specific tasks \cite{kollias2018training}, has led us to search for a better loss function than the categorical cross entropy. 

In fact, in the related face recognition field, it has been shown \cite{wen2016discriminative,liu2017sphereface} that categorical cross entropy loss is insufficient to acquire discriminating power for face classification. Several loss functions have been proposed for maximizing inter-class and minimizing intra-class variance. \cite{chopra2005learning,hoffer2015deep} propose multi-loss learning to increase
feature discriminating power. These, require thorough mining of pair/triplet samples, which is a time-consuming procedure. \cite{liu2017sphereface} projects the original Euclidean space of features to an angular space, introducing an angular margin for larger inter-class variance. \cite{wang2018cosface} directly adds a cosine margin penalty to the target logit, showing better performance than \cite{liu2017sphereface}. \cite{deng2018arcface} further improved the discriminative power of face recognition models, stabilising the training process. As these losses boosted face recognition models performance, in this work, we choose to adopt the ArcFace loss \cite{deng2018arcface} and adapt it for emotion recognition. To the best of our knowledge, this is the first time that such a loss designed for face recognition, is used in the context of emotion recognition. Our final contribution in this paper is the design of 2 networks trained with the ArcFace loss. After training them on Aff-Wild2, we re-trained them on each of the examined databases. Our results outperformed all state-of-the-art networks, illustrating: i) the richness of Aff-Wild2 (providing it with the ability to be used as robust prior for network pre-training) and ii) that the ArcFace loss can be used in the emotion recognition field, yielding state-of-the-art results. In fact, this is the very first proof of the effectiveness of additive angular margin in emotion recognition.

%


\begin{figure}
\centering
\scalebox{0.8}{
\begin{tabular}{c}
  \includegraphics[height=1.5cm]{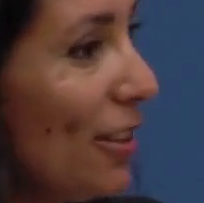}\includegraphics[height=1.5cm]{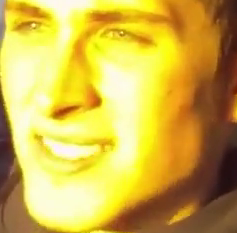}\includegraphics[height=1.5cm]{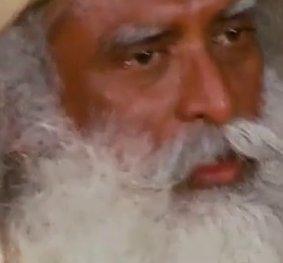}\includegraphics[height=1.5cm]{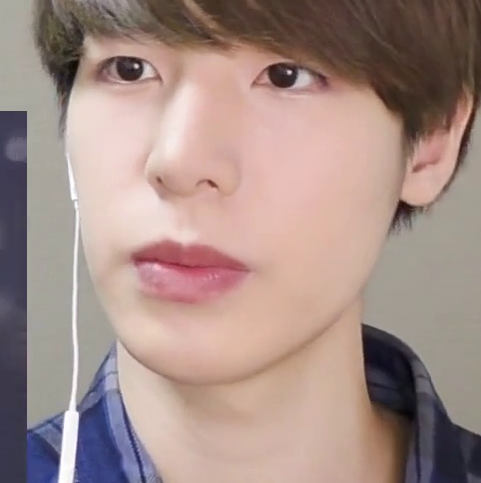}\includegraphics[height=1.5cm]{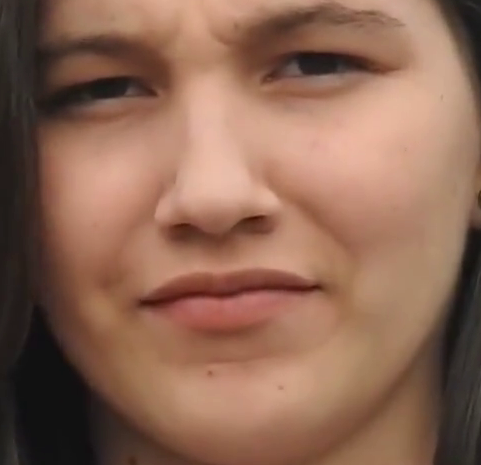}\includegraphics[height=1.5cm]{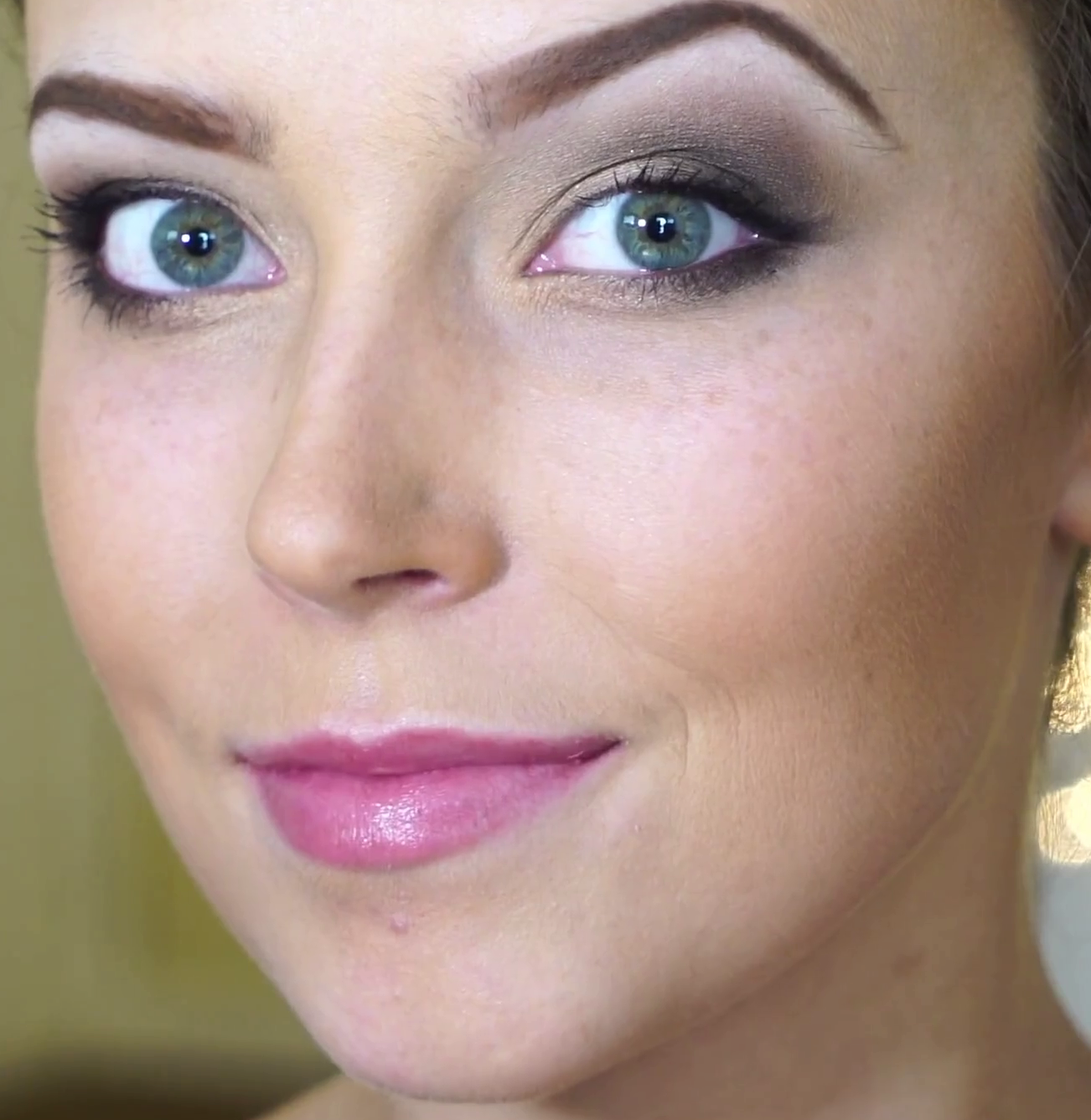}\includegraphics[height=1.5cm]{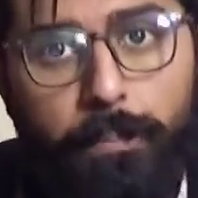}\includegraphics[height=1.5cm]{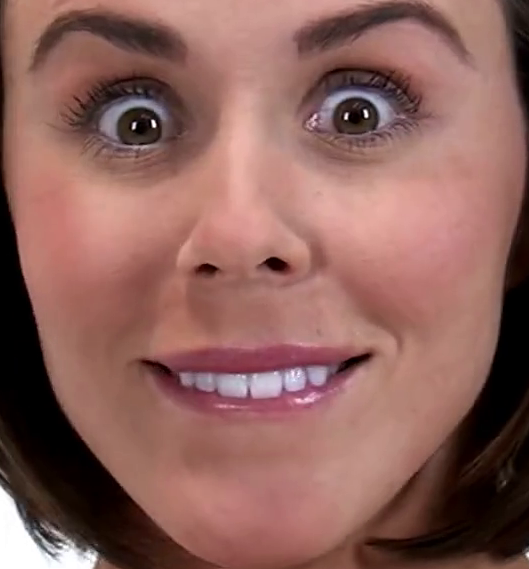}\includegraphics[height=1.5cm]{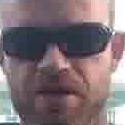}\includegraphics[height=1.5cm]{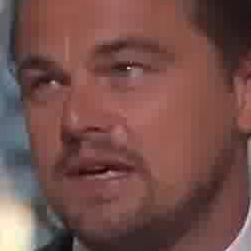}\\
\includegraphics[height=1.5cm]{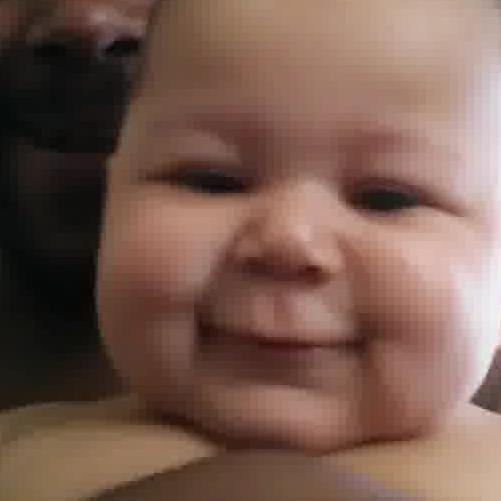}\includegraphics[height=1.5cm]{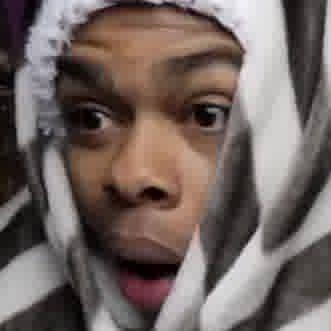}\includegraphics[height=1.5cm]{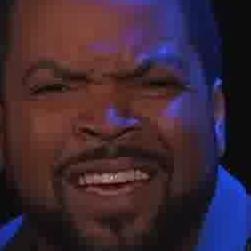}\includegraphics[height=1.5cm]{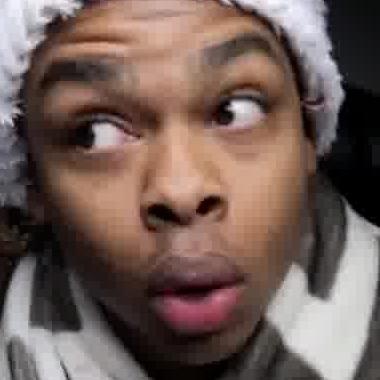}\includegraphics[height=1.5cm]{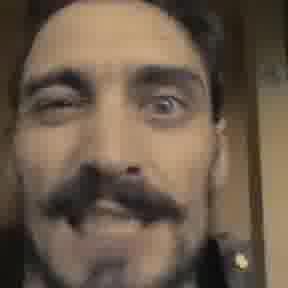}\includegraphics[height=1.5cm]{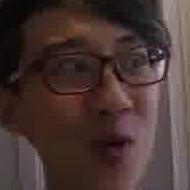}\includegraphics[height=1.5cm]{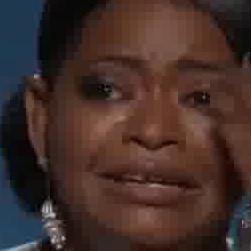}\includegraphics[height=1.5cm]{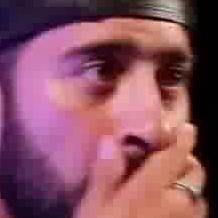}\includegraphics[height=1.5cm]{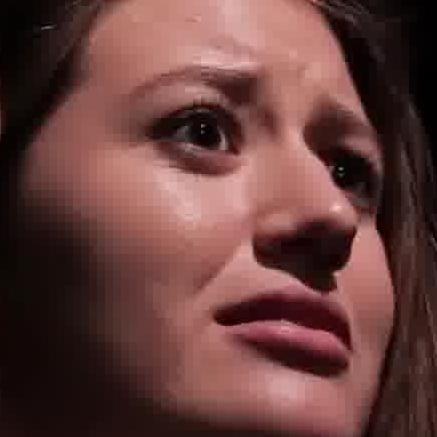}\includegraphics[height=1.5cm]{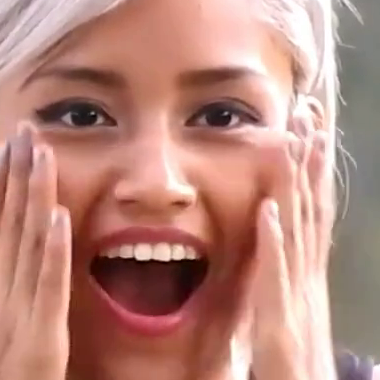} \\
\end{tabular}
}
\caption{Frames of Aff-Wild2, showing subjects of different ethnicities, age groups,  emotional states, head poses, illumination conditions and occlusions}
\label{frames_from_db}
\end{figure}

\section{The Aff-Wild2 database}
\begin{figure}[h]
\centering
\adjincludegraphics[height=3.3cm,width=7.8cm]{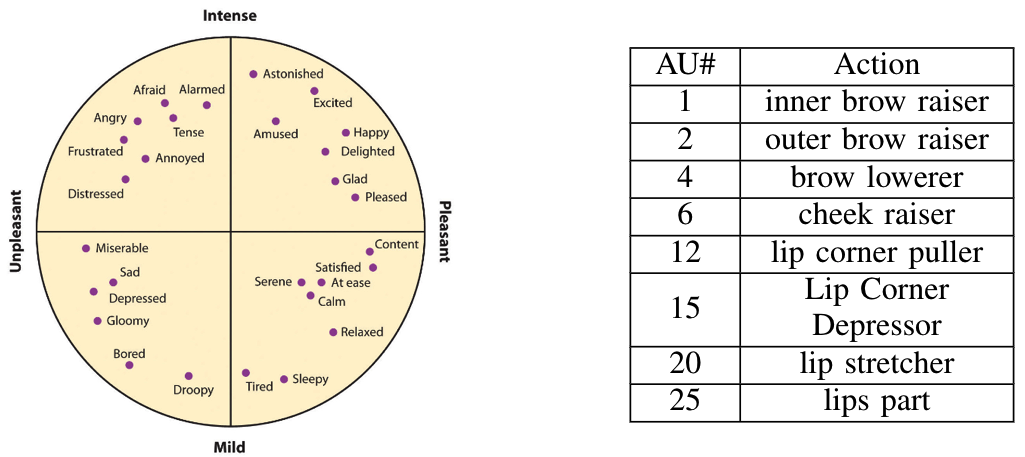}
\caption{The 2D Emotion Wheel (left); the AUs annotated in Aff-Wild2 (right) }
\label{2d-wheel-au}
\end{figure}

Aff-Wild2 is described next, presenting the new collected dataset and its properties, the generated partition sets, their distributions and the annotation procedure.  
\paragraph{Collected dataset and properties}
We extend the Aff-Wild database \cite{kollias2019deep,zafeiriou2017aff}, by collecting a new dataset consisting of \textit{260} YouTube videos, with \textit{1,413,000} 
frames and a total length of \textit{13} hours and \textit{5} minutes. 
The videos have been collected using the Youtube video sharing website. All of the collected videos are in MP4 format, with a frame rate of \textit{30}, provided under the CC licence. Keywords for retrieving the videos were selected from the 2-D Emotion Wheel, shown in Figure \ref{2d-wheel-au}.
The new videos have wide range in subjects': age (from babies to elderly people); ethnicity (caucasian/hispanic/latino/asian/black/african american); profession (e.g. actors, athletes, politicians, journalists); head pose; illumination conditions; occlussions; emotions. Figure \ref{frames_from_db} shows frames of Aff-Wild2 verifying the above described ranges.
These videos show subjects who: react on a surprise, on something that brings them happiness or fulfillment, on flirting or rejection, on important political issues, on funny or mean tweets; are stand-up comedians; give a really interesting speech in ceremonies; are taking an oral exam; are giving lectures on depression, or other serious disorders; are performing passive, boring, apathetic, intense activities,  etc.

Four experts annotated the new dataset in terms of valence and arousal, as in the case of Aff-Wild. 
We then concatenated the Aff-Wild database with the new dataset, forming Aff-Wild2. In total, Aff-Wild2 consists of \textit{558} videos with \textit{2,786,201} frames, showing both subtle and extreme human behaviours in real-world settings. The total number of subjects is \textit{458}; \textit{279} of which are males and \textit{179} females.

Two more tasks were implemented, in which we annotated parts of Aff-Wild2 with AUs and Exprs. 
In the first, three very experienced annotators annotated \textit{63} videos, with \textit{397,800}  frames and a total length of \textit{3} hours and \textit{41} mins, in terms of AUs \textit{1,2,4,6,12,15,20,25} - described in Figure \ref{2d-wheel-au}. These videos contain \textit{31} male and \textit{31} female subjects.
In the second, three experts annotated \textit{84} videos consisting of \textit{403,758} frames,  with a total length of \textit{3} hours and \textit{45} mins, in terms of the 7 basic expressions. The videos  show \textit{42}  male  and  \textit{42}  female subjects. 
Consequently, Aff-Wild2 contains 3 datasets (VA, AU, Expr); each contains annotations for a respective behavior task (preliminary work regarding the 3 sets can be found in \cite{kollias2018fg,kollias2018multi}).  Table \ref{attrs} summarizes the attributes and properties of the three annotated sets of Aff-Wild2. 

\begin{table}[h]
\caption{General Attributes of Aff-Wild2; in the VA set, top row refers to the new dataset, while bottom row refers to Aff-Wild}
\label{attrs}
\centering
\scalebox{0.75}{
\begin{tabular}{|c|c|c|c|c|c|}
\hline
Aff-Wild2 & \# frames  & \# videos & \# annotators  & Video Length  & Mean Resolution \\ 
\hhline{|=|=|=|=|=|=|}
VA set  & \begin{tabular}{@{}c@{}} $1,413,000$  \\ $1,373,201$ \end{tabular} &  \begin{tabular}{@{}c@{}} $260$ \\ $298$ \end{tabular} &   \begin{tabular}{@{}c@{}} $4$  \\ $8$ \end{tabular} &  \begin{tabular}{@{}c@{}}  $0.03-26.22$ mins  \\ $0.10-14.47$ mins \end{tabular} &  \begin{tabular}{@{}c@{}} $1450 \times 900 $  \\ $607 \times 359$ \end{tabular} \\
\hline
AU set & $397,800$ & $63$ & $3$ & $0.03-26.22$ mins & $1500 \times 900 $ \\
\hline
Expr set &  $403,758$ & $84$ & $3$ & $0.04-26.22$ mins & $ 1350 \times 800 $ \\    
\hline
\end{tabular}
}
\end{table}

\begin{figure}[h]
\centering
\adjincludegraphics[height=4.2cm,width=\linewidth]{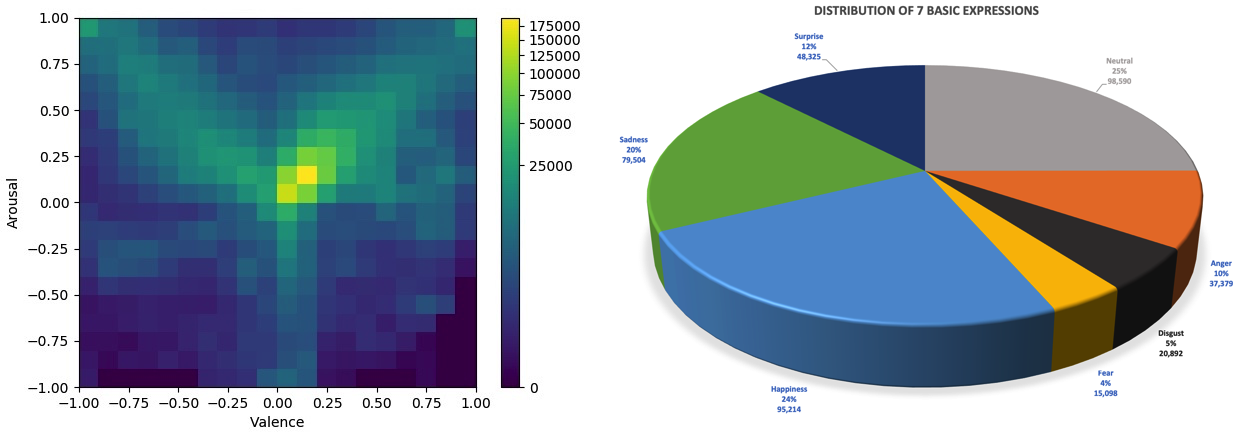}
\caption{2D VA Histogram of the new data added to Aff-Wild (left); Histogram of the seven basic expressions in Aff-Wild2 (right)}
\label{tab:frame_number_basic_expr}
\end{figure}

\paragraph{Partition Sets and Distributions} 
 \vskip-0.6cm
Each set (VA, AU, Expr) is split into three subsets: training, validation and test. Partitioning is done in a subject independent manner, in the sense that a person can  appear  only  in  one  of  those  three  subsets. In the VA set, the  resulting  training, validation and test subsets consist of \textit{350}, \textit{70} and \textit{138} videos 
respectively. In the AU set, the  respective  subsets consist of \textit{42}, \textit{7} and  \textit{14} videos 
respectively. In the Expr set, the  corresponding subsets consist of \textit{51}, \textit{11} and \textit{22} videos 
respectively.
Figure \ref{tab:frame_number_basic_expr} shows the 2D VA histogram of the new dataset, which was added to Aff-Wild.
Figure \ref{tab:frame_number_basic_expr} shows the distribution of the seven emotion categories in Aff-Wild2. 
Table \ref{tab:frame_number_action_unit} shows the distribution of the activated AUs. 
We note that the Expr Set of images can be extended to also contain AU annotations, according to Table 1 of \cite{du2014compound}. 
However, this is out of the scope of the current paper.



\begin{table}[h]
    \centering
        \caption{Distribution of AU annotations in Aff-Wild2}
    \label{tab:frame_number_action_unit}
\scalebox{0.62}{    
\begin{tabular}{|c|c|c|c|c|c|c|c|c|}
\hline
  Action Unit \# & AU 1 & AU 2 & AU 4 & AU 6 & AU 12 & AU 15 & AU 20 & AU 25 \\
  \hline    
  \begin{tabular}{@{}c@{}}Total Number  of Activated AUs \end{tabular}  & \begin{tabular}{@{}c@{}} 86,677 \\ 43.9 \%  \end{tabular} & \begin{tabular}{@{}c@{}} 4,166 \\ 2.1\%  \end{tabular} & \begin{tabular}{@{}c@{}} 56,327  \\ 28.5\% \end{tabular} & \begin{tabular}{@{}c@{}} 25,226 \\ 12.8\% \end{tabular} & \begin{tabular}{@{}c@{}} 35,675 \\ 18.1\% \end{tabular} & \begin{tabular}{@{}c@{}} 3,340 \\ 1.7\% \end{tabular} & \begin{tabular}{@{}c@{}} 5,695 \\ 2.9\% \end{tabular} & \begin{tabular}{@{}c@{}} 9,048 \\  4.6\%  \end{tabular}\\
  \hline
\end{tabular}
}
\end{table}

\paragraph{Annotation} 
 \vskip-0.3cm
Four experts have performed the VA set annotation, using the method proposed in \cite{cowie2000feeltrace}. 
Valence and arousal values range continuously in [-1,1]. The final label values are the mean of those four annotations. The mean inter-annotation correlation is 0.63 for valence and 0.60 for arousal. For the AU set, three experts have performed the annotation. For the Expr set, three more experts performed the annotation.  In both cases, agreement between the annotators has not always been 100\%. We only kept the annotations, on which all experts agree.

\section{Proposed Methods}
\vskip-0.15cm
\noindent Two pre-processing steps, on the visual and audio modalities, have been applied to generate the input data for DNN based emotion analysis. We developed the following deep network architectures for emotion recognition: i) CNN, single-/multi- task; ii) CNN-RNN multi-task; iii) CNN-RNN multi-modal (A/V) and multi-task; iv) new ArcFace networks with respective loss function, as described below.  

\paragraph{Visual Modality Pre-Processing}  
The SSH detector \cite{najibi2017ssh} based on the ResNet and trained on the WiderFace dataset \cite{yang2016wider} was used to extract face bounding boxes from all images. Also, 5 facial landmarks (two eyes, nose and two mouth corners) were extracted and used to perform similarity transformation (for face alignment). After that we obtain the cropped faces which are then resized to dimension $96 \times 96 \times 3$. The pixel intensities are normalized to take values in [-1,1].

\paragraph{Audio Modality Pre-Processing}  
The audio signal (mono) is sampled at $44,100$Hz. Then spectrograms are extracted; spectrogram frames are computed over a $33$ms window with $11$ms overlap. The resulting intensity values are normalized in [-1,1] to be consistent with the visual modality.

\paragraph{CNN Single- \& Multi-Task} 
We employ 3 state-of-the-art networks, SphereFace-20 \cite{liu2017sphereface},  VGGFace \cite{parkhi2015deep}, and Inception ResNet \cite{szegedy2017inception} (denoted as Inc.ResNet). We train these networks to perform one behavior task (VA estimation, AU detection, or Expr classification), or jointly  perform all 3 tasks.
We call the multi-task VGGFACE network, MT-VGG.
The predictions for all tasks are pooled from the same feature space.

\paragraph{CNN-RNN Multi-Task} 
As shown in the experimental section, MT-VGG has the best performance; thus we construct a CNN-RNN multi-task network, based on MT-VGG. In more detail, a 2-layer GRU with 128 cells each is stacked on top of the first fc layer of MT-VGG for capturing the temporal dynamics; the output layer is on top of the GRU. We call this network MT-VGG-RNN.

\begin{figure}[h]
\centering
 \includegraphics[height=3cm,width=8.5cm]{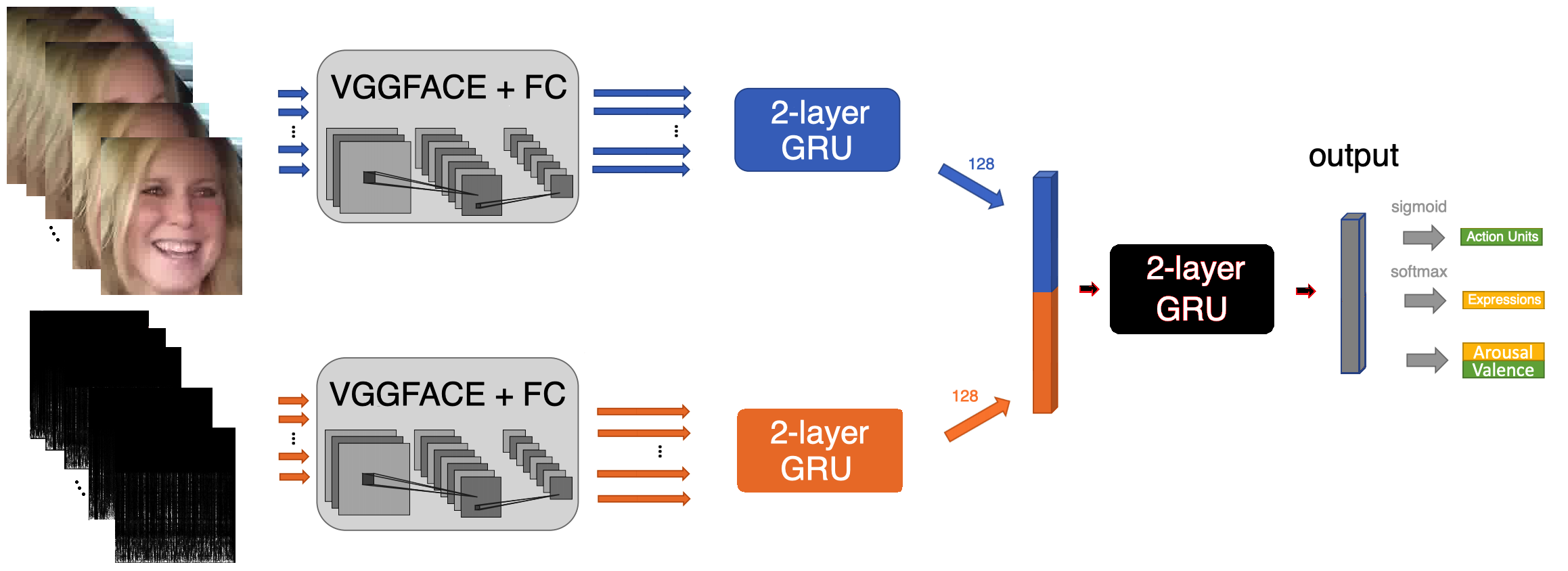} 
\caption{A/V-MT-VGG-RNN: the Multi-Modal and Multi-Task developed model}
\label{multi-modal}
\end{figure}

\paragraph{CNN-RNN Multi-Modal (A/V) \& Multi-Task} 
To handle both video and audio modalities, we use a feature level fusion strategy in our  developed deep learning model, that we illustrate in Figure \ref{multi-modal}. 
This model consists of two identical streams that extract features directly from raw input images and spectrograms, respectively. Each stream consists of a MT-VGG-RNN, described above, without the output layer. The features from the two streams are concatenated, forming a 256-dimensional feature vector that is passed through a 2-layer GRU layer with 128 units in each layer, in order to fuse the information of the audio and visual streams. The output layer follows on top of it. We call this network A/V-MT-VGG-RNN.

\paragraph{Standard Loss Functions} 
The objective function minimized during training of the multi-task networks is the sum of the individual task losses:
\vspace*{-4px}
\begin{align}
\mathcal{L}_{CCE} &= \mathbb{E}[-\text{log } e^{p_p} + \text{log } \sum\nolimits_{i=1}^{7}{e^{p_i}} ]
\label{eq:cce}\\
\mathcal{L}_{BCE} &= \mathbb{E}[ -\sum\nolimits_{i=1}^{17}{(t_i \cdot \text{log } p_i + (1-t_i) \cdot \text{log } (1 -p_i))}] 
\label{eq:bce}\\
\mathcal{L}_{CCC} &= 1- 0.5 \cdot (\rho_a + \rho_v) \text{, with } \rho_{a,v} =  2 s_{xy} \div [s_x^2 + s_y^2 + (\bar{x} - \bar{y})^2] 
\label{eq:ccc}
\end{align}
where $\mathcal{L}_{CCE}$ is the categorical cross entropy loss, $\mathcal{L}_{BCE}$ is the binary cross entropy loss,  $p_p$  is the prediction of positive class, $p_i$ is the prediction of $AU_i$,  $t_i$ $\in \{0,1\}$ is the label of $AU_i$, $\rho_{a,v}$ is the Concordance Correlation Coefficient (CCC) of arousal/valence, $s_x$ and $s_y$ are the variances of arousal/valence labels and predicted values respectively and $s_{xy}$ is the corresponding covariance value. 

\paragraph{ArcFace Loss Function \& Networks}
Next, we focus on Expr recognition and  introduce a new loss function. 
The softmax cross-entropy loss is modified as follows:
\begin{align}
\resizebox{.94\hsize}{!}{$\displaystyle{
\mathcal{L} = \frac{-1}{N} \sum_{i=1}^{N} {\log\frac{e^{W_{y_i}^Tx_i}}{\sum_{j=1}^{7}{e^{W_{j}^Tx_i}}} } = \frac{-1}{N} \sum_{i=1}^{N} {\text{log}\frac{e^{\norm{W_{y_i}} \cdot \norm{x_i} \cdot  \cos{\theta_{y_i}} }}{\sum_{j=1}^{7}{e^{\norm{W_j} \cdot \norm{x_i} \cdot  \cos{\theta_{j}}} }} }   \numeq {\norm{x_i}=s,} \frac{-1}{N} \sum_{i=1}^{N} {\text{log}\frac{e^{s\cdot \cos{\theta_{y_i}}  }}{\sum_{j=1}^{7}{e^{s\cdot \cos{\theta_{j} }} } } }
}$
}
\label{eq:mt2}
\end{align}
\noindent where the embedding feature $x_i \in \mathcal{R}^d$  denotes the deep feature of the \textit{i}-th sample belonging to the $y_i$-th class, $W_j \in \mathcal{R}^d$ denotes the \textit{j}-th column of the weight $W \in \mathcal{R}^{d \times 7}$, \textit{N} is the batch size, $\theta_j$ is the angle between weight $W_j$ and feature $x_i$, $\norm{W_j}$ is fixed to \textit{$1$} by \textit{$l_2$} normalization, $\norm{x_i}$ is fixed by \textit{$l_2$} normalization and re-scaled to \textit{$s$}.

From eq.\ref{eq:mt2}, it can be seen that the embedding features are distributed around each feature centre on the hypersphere. In our case, we adopt the ArcFace loss, where an angular margin penalty \textit{$m$} between $x_i$  and $W_{y_i}$ is added to simultaneously
enhance the intra-class compactness and inter-class discrepancy (eq.\ref{eq:mt2}: $\theta_{y_i} \xrightarrow{} \theta_{y_i} + m$).  \textit{$m$} is equal to the geodesic distance margin penalty in the normalised hypersphere. We refer the interested reader to \cite{deng2018arcface} for more details and explanation of this loss.

\begin{figure}[h]
\centering
\scalebox{0.92}{
\adjincludegraphics[height=3cm,width=13.7cm]{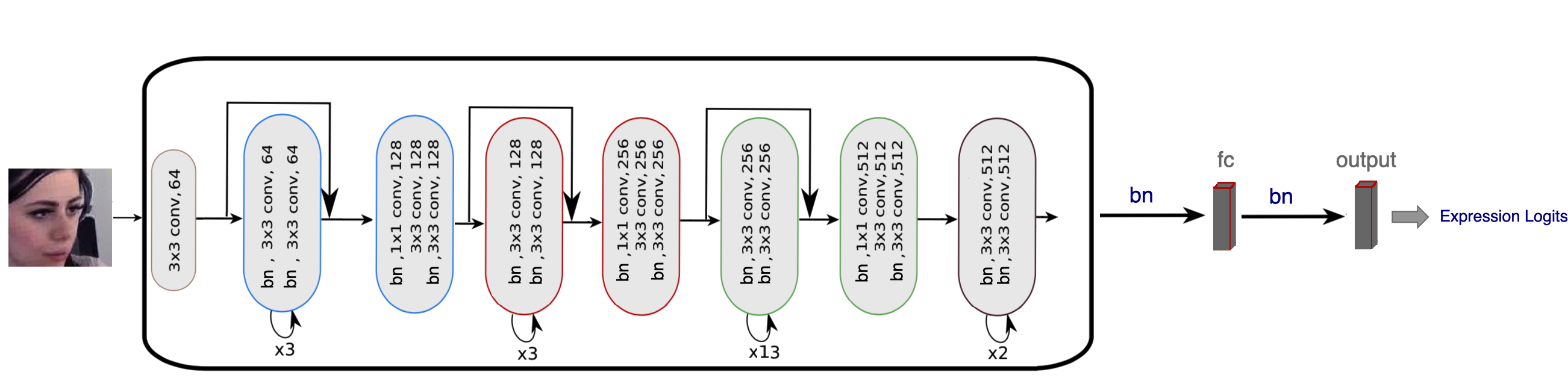}
}
\caption{The MT-ArcRes network that has been trained with the ArcFace loss}
\label{ArcResnet}
\end{figure}

Next, we develop two networks to account for this loss. The first CNN architecture, called Multi-Task-ArcFace-Residual (MT-ArcRes) uses residual units and is depicted in Fig.\ref{ArcResnet}; 'bn' stands for batch normalization, the convolution layer is in the format: filter height $\times$ filter width conv., number of output feature maps; the stride is equal to 2, everywhere; the fc layer is the embedding layer; the output layer provides the seven expresion class logits ($W_j^T x_i$, $j=1..7$). The second network is called  Multi-Task-Arcface-VGG (MT-ArcVGG); the difference with MT-ArcRes is that the rectangular area in the Figure contains VGGFace's  layers. 


\section{Experimental Study}
 \vskip-0.2cm
\noindent The experimental study consists of two parts. In the first, we train MT CNNs, CNN-RNNs \& Multi-Modal CNN-RNNs, for VA, AU and Expr Recognition on the Aff-Wild2; then, we test the networks on 10 different databases, showing that Aff-Wild2 and the Multi-Task networks provide the best pre-trained framework for a large variety of emotion recognition settings. In the second, focusing on expression recognition, we first train ArcFace networks with Aff-Wild2 and then re-train them with each expression database; we evaluate them, achieving state-of-the-art performance.

\paragraph{Implementation Details \& Settings}
Specific details about hyperparameters of the developed architectures can be found in Table \ref{hyper}.  All experiments in this paper are implemented in TensorFlow, on a Tesla V100 32GB GPU, using Adam optimizer (with default values) or SGD with momentum (0.9) in the ArcFace Networks' case. Additional details follow:

\noindent I) \underline{\textit{CNN Single- \& Multi-Task}}: The networks have first been pre-trained for VA estimation on the Aff-Wild database, then the output layer is discarded and substituted by a new one for single- or multi-task, depending on the network type. Then they are trained end-to-end on Aff-Wild2. 

\noindent II) \underline{\textit{CNN-RNN Multi-Task}}: The CNN part is initialized with the weights of the CNN MT-VGG. Then the whole architecture is trained end-to-end on Aff-Wild2. 

\noindent III) \underline{\textit{CNN-RNN Multi-Modal (A/V) \& Multi-Task}}: Training is divided in two phases: first the audio/visual streams are trained independently and then the audiovisual network is trained end-to-end. To train each stream individually, we follow the same procedure as in the CNN-RNN Multi-Task case. Once the single streams are trained, they are used for initializing the corresponding streams in the multi-stream architecture. Finally, the entire audiovisual network is trained end-to-end. 

\noindent IV) \underline{\textit{ArcFace Networks}}: Both networks are first trained on Aff-Wild2. Then, they are re-trained end-to-end on each of the examined databases. 
During testing 
we keep the feature embedding layer, discarding the output layer. For all training images, we extract features from the embedding layer and split them in 7 clusters. Then, for each test image, we compute its distance (based on cosine similarity) from all cluster centers and assign it to the center for which this distance is minimum. 

\begin{table}[h]
\caption{Network Configurations: ST= Single Task, MT=Multi-Task} 
\label{hyper}
\centering
\scalebox{0.58}{
\begin{tabular}{ |c||c|c|c|c|}
 \hline
\multicolumn{1}{|c||}{} & \multicolumn{1}{c|}{\begin{tabular}{@{}c@{}} ST- \& MT- CNN \end{tabular}} & 
\multicolumn{1}{c|}{ MT-CNN-RNN }  & \multicolumn{1}{c|}{A/V-MT-CNN-RNN} & \multicolumn{1}{c|}{MT-ArcRes / MT-ArcVGG } \\
  \hhline{=:=:=:=:=}
 learning rate & $[10^{-4},10^{-5}], best: 10^{-4} $  & $[10^{-4},10^{-6}], best: 10^{-5}$  & $[10^{-3},10^{-6}], best: 10^{-5}$ & $[10^{-4},10^{-5}], best: 10^{-4} $   \\
\hline
 batch size/seq.length  & 256 / -  & 10 / 90 & 5 / 90  &  300 / -  \\
\hline
parameters  & dropout=0.4  & dropout=0.4 & dropout=0.4  &  \begin{tabular}{@{}c@{}} dropout=0.4, $d \in \{32,512\}, s \in \{32,64\},$ \\ $ m \in \{0.1,0.5,1,1.5,2,2.5,3\}, best: 0.1/1  $ \\   \end{tabular}   \\
\hline
\end{tabular}
}
\end{table}

\paragraph{Databases} 
Table \ref{databases} shows the databases used in our experiments along with their properties. BP4DS and BP4D+ datasets correspond to the ones used in the FERA 2015 \cite{valstar2015fera} and 2017 \cite{valstar2017fera} Challenges, respectively. All databases are in-the-wild, apart from DISFA, BP4DS, BP4D+, which are spontaneous. Let us note that for the AffectNet, BP4DS and BP4D+ databases, the test set is not released; thus we report the performances on the validation set, which we use for testing.

\begin{table}[h]
\caption{Properties of Databases used in our Experiments} 
\label{databases}
\centering
\scalebox{0.55}{
\begin{tabular}{ |c||c|c|c|c|c|c|c|c|c| }
 \hline
\multicolumn{1}{|c||}{\begin{tabular}{@{}c@{}} Databases  \end{tabular}} & \multicolumn{1}{c|}{\begin{tabular}{@{}c@{}}  AFEW-VA \cite{kossaifi2017afew} \end{tabular}} & \multicolumn{1}{c|}{  AffectNet }  & \multicolumn{1}{c|}{RAF-DB \cite{li2017reliable}} & \multicolumn{1}{c|}{FER2013 \cite{goodfellow2013challenges}} & \multicolumn{1}{c|}{IMFDB\cite{setty2013indian}} & \multicolumn{1}{c|}{Emotionet} & \multicolumn{1}{c|}{DISFA \cite{mavadati2013disfa}} & \multicolumn{1}{c|}{BP4DS \cite{zhang14bp4d}} & \multicolumn{1}{c|}{BP4D+ \cite{zhang2016multimodal}}\\
  \hhline{=:=:=:=:=:=:=:=:=:=}
  Model of Affect & VA & VA, Expr & Expr &  Expr   &  Expr &  AUs &  AUs &  AUs &  AUs \\
\hline
  \# of videos & 600 & - & - & - & - & - & 54 & 1,640 & 5,463 \\
\hline
  \# of frames & 30,050 &  450,000 & 15,200 & 35,887 & 34,512 & 50,000 & 261,630 & 222,573 & 967,570 \\
\hline
\end{tabular}
}
\end{table}

\paragraph{Evaluation Metrics}
CCC, defined in eq.\ref{eq:ccc} is used for VA estimation, as it has been the evaluation criterion in all related Challenges \cite{kollias2019deep,ringeval2017avec}. The usual F1 score is adopted for evaluation of AU detection and Expr classification. Exceptions are the RAF-DB and FER2013, in which the mean diagonal value of the confusion matrix and the accuracy metric, respectively, are the default performance measures.

\paragraph{Results on static databases for VA \& Expr Recognition}
Table \ref{comparison_static} presents the results of different CNN Single- and Multi-task (ST- and MT-) networks in a cross-database setting (networks are trained on Aff-Wild2 and tested on AffectNet, RAF-DB, FER2013 and IMFDB). The MT-VGG has the best performance for both VA and Expr recognition. In Table \ref{comparison_static}, we also compare MT-VGG's performance with that of the state-of-the-art in each of the tested databases (the results shown are taken from the respective papers). It can be seen that MT-VGG beats the state-of-the-art in all databases, illustrating the excellent cross-performance of the generated framework.  Only, in expression recognition in AffectNet, the obtained performance is lower to the state-of-the-art. 


\skip -0.6cm
\begin{table}[h]
\skip -0.6cm
\caption{Cross-database evaluation (models trained on Aff-Wild2 and tested on other databases) 
for VA and Expr on static databases: 
VA evaluation is shown as ($CCC_V$-$CCC_A$); single values correspond to expressions' performance metrics} 
\label{comparison_static}
\centering
\scalebox{0.6}{
\begin{tabular}{ |c||c|c|c|c|c|c|c|c|c| }
 \hline
\multicolumn{1}{|c||}{\begin{tabular}{@{}c@{}} Databases  \end{tabular}} &  \multicolumn{1}{c|}{  MT-VGG }  & \multicolumn{1}{c|}{\begin{tabular}{@{}c@{}} ST-VGG\end{tabular}} & \multicolumn{1}{c|}{\begin{tabular}{@{}c@{}} MT- \\ SphereFace\end{tabular}} & \multicolumn{1}{c|}{\begin{tabular}{@{}c@{}} ST- \\ SphereFace\end{tabular}} &  \multicolumn{1}{c|}{\begin{tabular}{@{}c@{}} MT- Inc. \\ ResNet\end{tabular}} &  \multicolumn{1}{c|}{\begin{tabular}{@{}c@{}} ST- Inc. \\ ResNet\end{tabular}}  & \multicolumn{1}{c|}{AlexNet \cite{mollahosseini2017affectnet}}   & \multicolumn{1}{c|}{VGGFACE\cite{li2017reliable}} 
&
\multicolumn{1}{c|}{VGG\cite{georgescu2018local}}      \\
\hhline{=:=:=:=:=:=:=:=:=:=}
FER2013 & \textbf{0.76}  & 0.73  &  0.72  & 0.72  &  0.74  & 0.71  & -    & - & 0.75  \\
\hline
RAF-DB & \textbf{0.61}  & 0.57  & 0.53   & 0.52   & 0.57 & 0.55  & -  & 0.58   & - \\
\hline
IMFDB & \textbf{0.42}  & 0.39  &  0.39  & 0.38  &  0.4  & 0.39  & -   & -  & -  \\
\hline
AffectNet & \begin{tabular}{@{}c@{}} (\textbf{0.61}-\textbf{0.46})  \\ \textit{0.54}  \end{tabular}  & \begin{tabular}{@{}c@{}} (0.51-0.42) \\ 0.52   \end{tabular}   &  \begin{tabular}{@{}c@{}} (0.5-0.43) \\0.5  \end{tabular}   & \begin{tabular}{@{}c@{}} (0.5-0.4) \\ 0.51  \end{tabular}    &  \begin{tabular}{@{}c@{}} (0.52-0.45) \\ 0.52  \end{tabular}   & \begin{tabular}{@{}c@{}} (0.5-0.42) \\ 0.51  \end{tabular} & \begin{tabular}{@{}c@{}} (0.6-0.34) \\ \textbf{0.58}  \end{tabular} & -  & -  \\
  \hline
\end{tabular}
}
\end{table}

\paragraph{Results on video databases for VA \& Expr Recognition}
 \vskip-0.45cm
Table \ref{comparison_video} presents the results of CNN, CNN-RNN multi-task,  single- and multi-modal networks in a cross-database setting, testing on Aff-Wild, Aff-Wild2 and AFEW-VA databases. It can be seen that the MT-VGG-RNN (trained on the visual modality) displays a better performance than the MT-VGG, for VA and Expr Recognition, in all databases. Moreover,  MT-VGG-RNN performs best for valence estimation when trained with the visual modality, whereas performs best for arousal when trained with the audio modality. This is because audio tends to have thematic constancy. Consider, for example, two fight sequences in a movie, one being a flashy fight scene and the other a one-sided fight with a person being injured. In both cases, arousal can be high due to loud and pronounced music, but valence will be positive in the former and negative in the latter sequence. 
It can also be seen that the A/V-MT-VGG-RNN outperforms the MT-VGG-RNN, illustrating that the A/V combination improves network performance, in valence, arousal and expression estimation. 
Table \ref{comparison_video} compares the performance of the proposed networks with the state-of-the-art in the examined databases 
. It is evident that MT-VGG and MT-VGG-RNN outperform the respective state-of-the-art.

\begin{table}[h]
\caption{Cross-database evaluation  for VA and Expr  
on video databases: VA evaluation is shown as ($CCC_V$-$CCC_A$); single values correspond to F1 score} 
\label{comparison_video}
\centering
\scalebox{0.7}{
\begin{tabular}{ |c||c|c|c|c|c|c| }
 \hline
\multicolumn{1}{|c||}{\begin{tabular}{@{}c@{}} Databases  \end{tabular}} &  \multicolumn{1}{c|}{  MT-VGG }  & \multicolumn{1}{c|}{\begin{tabular}{@{}c@{}} MT-VGG-RNN \\ visual modality \end{tabular}} & \multicolumn{1}{c|}{\begin{tabular}{@{}c@{}} MT-VGG-RNN \\  audio modality \end{tabular}}  &  \multicolumn{1}{c|}{\begin{tabular}{@{}c@{}} A/V-MT-VGG-RNN \end{tabular}} &  \multicolumn{1}{c|}{\begin{tabular}{@{}c@{}} best CNN \\  \cite{kollias,kollias2019deep}\end{tabular}}  & \multicolumn{1}{c|}{\begin{tabular}{@{}c@{}}AffWildNet \\  \cite{kollias,kollias2019deep}\end{tabular}}      \\
  \hhline{=:=:=:=:=:=:=}
  Aff-Wild &  (0.56-0.35)     & (0.60-0.45)  &  (0.51-0.47)      &  \textbf{(0.62-0.49)}   &  (0.51-0.33)  &  (0.57-0.43)     \\
  \hline
  Aff-Wild2 & \begin{tabular}{@{}c@{}}    (0.38-0.3) \\ 0.4 \end{tabular} &
  \begin{tabular}{@{}c@{}}    (0.40-0.33) \\ 0.43 \end{tabular}  & 
 \begin{tabular}{@{}c@{}}    (0.34-0.36) \\ 0.43 \end{tabular}  &
   \begin{tabular}{@{}c@{}}    \textbf{(0.42-0.38)} \\ \textbf{0.46} \end{tabular}  & \begin{tabular}{@{}c@{}}    (0.33-0.25) \\ - \end{tabular}  & \begin{tabular}{@{}c@{}}    (0.35-0.28) \\ - \end{tabular}     \\
  \hline
  AFEW-VA & (0.58-0.53)  & \textbf{(0.6-0.6)} & -    & -  & (0.49-0.52) & (0.52-0.56)      \\
  \hline
\end{tabular}
}
\end{table}

\paragraph{Results for AU Detection} 
 \vskip-0.35cm
Table \ref{comparison_au} compares the performance between MT-VGG and state-of-the-art networks in a cross-database setting among Emotionet, DISFA, BP4DS  and BP4D+; all reported results are for the common AUs between the testing database and Aff-Wild2. It is clear that MT-VGG outperforms the winner \cite{ding2017facial} of Emotionet 2017 Challenge, the baseline \cite{valstar2015fera} and the winner \cite{yuce2015discriminant} of FERA 2015, the baseline \cite{valstar2017fera} of FERA 2017 and the fine-tuned VGG (FVGG) and R-TI method of \cite{li2017action}. Apart from Emotionet, in all other cases, there is a boost in performance. MT-VGG displays a slightly worse performance than the winner \cite{tang2017view} of FERA 2017.





\begin{table}[ht]
\caption{Cross-database evaluation for AU Detection: evaluation metric is  F1 score}
\label{comparison_au}
\centering
\scalebox{0.75}{
\begin{tabular}{ |c||c|c|c|c|c|c|c|c|c| }
 \hline
\multicolumn{1}{|c||}{\begin{tabular}{@{}c@{}} Databases \end{tabular}} 
&\multicolumn{1}{c|}{MT-VGG} & \multicolumn{1}{c|}{MT-VGG-RNN}  
&\multicolumn{1}{c|}{\cite{ding2017facial}}
&\multicolumn{1}{c|}{\cite{yuce2015discriminant}}
&\multicolumn{1}{c|}{\cite{valstar2015fera}} &\multicolumn{1}{c|}{\cite{tang2017view}} &\multicolumn{1}{c|}{\cite{valstar2017fera}} 
&\multicolumn{1}{c|}{FVGG\cite{li2017action}}
&\multicolumn{1}{c|}{R-T1\cite{li2017action}}
\\
  \hhline{=:=:=:=:=:=:=:=:=:=} 
Aff-Wild2 &  0.42 &  \textbf{0.44}  & -  & - & - & - & - & -  & - \\
\hline
Emotionet &  \textbf{0.52}  & -   &  0.51  &  - & - & - & - & - & - \\
\hline
DISFA &  \textbf{0.61}  & - & - & - & - & - & - &  0.52  &  0.60    \\
 \hline
BP4DS &  \textbf{0.66} & - & - &  0.54   & 0.53   & - & - & - & - \\ 
\hline
BP4D+ &  \textit{0.49}  & - & - & - &  - &  \textbf{0.51}  &  0.34 & - & - \\
\hline 
\end{tabular}
}
\end{table}

From the above Tables, it is evident that Aff-Wild2 constitutes a very rich database for deep network training and further testing on very different and diverse emotion databases; the presented cross-database results validate our network developments.

\paragraph{Results with ArcFace Loss for Expr Recognition}
Table \ref{comparison_arcface} presents a performance comparison between: i) MT-VGG (trained on Aff-Wild2), ii) a fine-tuned MT-VGG (FT-MT-VGG; pre-trained on Aff-Wild2, then re-trained on each of the examined databases), iii) the two networks trained with the ArcFace loss (MT-ArcRes, MT-ArcVGG) on Aff-Wild2 and re-trained on each of the examined databases, iv) the state-of-the-art in these databases (whose results are taken from the respective papers). The FT-MT-VGG outperforms the state-of-the-art in all databases, apart from RAF-DB, where DLP-CNN performs better, however, trained with a locality preserving loss function. Table 8 also compares this network's performance to the performance of 
MT-ArcRes and  MT-ArcVGG networks, trained with the ArcFace loss function. These networks outperform all others, including DLP-CNN.


\begin{table}[h]
\caption{Retrained multi-task networks with ArcFace loss, for expression recognition } 
\label{comparison_arcface}
\centering
\scalebox{0.7}{
\begin{tabular}{ |c||c|c|c|c|c|c|c|}
 \hline
\multicolumn{1}{|c||}{\begin{tabular}{@{}c@{}} Databases  \end{tabular}} 
&  \multicolumn{1}{c|}{MT-ArcRes} 
&  \multicolumn{1}{c|}{MT-ArcVGG} 
& \multicolumn{1}{c|}{FT-MT-VGG}
& \multicolumn{1}{c|}{MT-VGG}
& \multicolumn{1}{c|}{AlexNet \cite{mollahosseini2017affectnet}} 
& \multicolumn{1}{c|}{DLP-CNN \cite{li2017reliable}} 
& \multicolumn{1}{c|}{VGG\cite{georgescu2018local}}
\\
 \hhline{=:=:=:=:=:=:=:=}
 AffectNet & \textbf{0.63} & 0.62 & 0.59 & 0.54  & 0.58  & -   & -\\
\hline
 RAF-DB &  0.75 & \textbf{0.76}   & 0.71  & 0.61  & - & 0.74   & - \\
\hline
IMFDB  & 0.55&  \textbf{0.56}   &0.51  & 0.42 &- &- &  - \\
\hline
 FER2013 & \textbf{0.8} & 0.79    & 0.78  & 0.76  & - & -  & 0.75 \\
\hline
\end{tabular}
}
\end{table}


\section{Conclusions}
 \vskip-0.3cm
\noindent In this paper, we present the first, largest, in-the-wild, A/V database, called Aff-Wild2, that is annotated for VA, AUs and Exprs. We build and train multi-task and multi-modal CNNs and CNN-RNNs on Aff-Wild2 and test their performances on 10 databases, beating the state-of-the-art. We further train two new networks on Aff-Wild2, adopting the ArcFace loss function, and then re-train them on a variety of expression databases; the results improve the existing state-of-the-art.

\paragraph{Acknowledgements} We would like to thank Viktoriia Sharmanska for our fruitful conversations during preparation of this work. The work of S. Zafeiriou has been partially funded by the EPSRC Fellowship Deform (EP/S010203/1). The work of Dimitrios Kollias was funded by a Teaching Fellowship of Imperial College London.

\bibliography{egbib}
\end{document}